%% file: arxiv.tex
\documentclass[10pt,twocolumn,letterpaper]{article}

\usepackage[pagenumbers]{cvpr} %

\usepackage[accsupp]{axessibility}  %
\usepackage{graphicx}
\usepackage{amsmath}
\usepackage{amssymb}
\usepackage{booktabs}
\usepackage{multirow}
\usepackage{xcolor} 
\usepackage{soul}
\usepackage{pifont}%
\newcommand{\cmark}{\ding{51}}%
\newcommand{\xmark}{\ding{55}}%
\usepackage{makecell}
\usepackage{setspace}

\usepackage[pagebackref,breaklinks,colorlinks]{hyperref}

\usepackage[capitalize]{cleveref}
\crefname{section}{Sec.}{Secs.}
\Crefname{section}{Section}{Sections}
\Crefname{table}{Table}{Tables}
\crefname{table}{Tab.}{Tabs.}

\newcommand{\model}{Vid2Seq}

\def\cvprPaperID{386} %
\def\confName{CVPR}
\def\confYear{2023}

\begin{document}

\title{\model{}: Large-Scale Pretraining of a Visual Language Model\\ for Dense Video Captioning
\vspace{-0.5cm}}

\author{Antoine Yang$^\dag$\footnotemark[1] \;\; 
Arsha Nagrani$^\S$ \;\; 
Paul Hongsuck Seo$^\S$ \;\; 
Antoine Miech$^\sharp$ \;\; \\
Jordi Pont-Tuset$^\S$\;\; 
Ivan Laptev$^\dag$\;\; 
Josef Sivic$^\P$\;\; 
Cordelia Schmid$^\S$ \\
\small{
$^\S$Google Research \quad
$^\dag$Inria Paris and D\'{e}partement d'informatique de l'ENS, CNRS, PSL Research University} \\
\small{
$^\sharp$ DeepMind
\quad $^\P$Czech Institute of Informatics, Robotics and Cybernetics at the Czech Technical University in Prague}
\\
\small{\url{https://antoyang.github.io/vid2seq.html}}
}

\twocolumn[{%
\renewcommand\twocolumn[1][]{#1}%
\vspace{-0.5cm}
\maketitle
\pagestyle{plain}

\vspace{-3em}
\begin{center}
\includegraphics[clip, trim=0cm 11cm 9cm 0cm, width=1.0\linewidth]{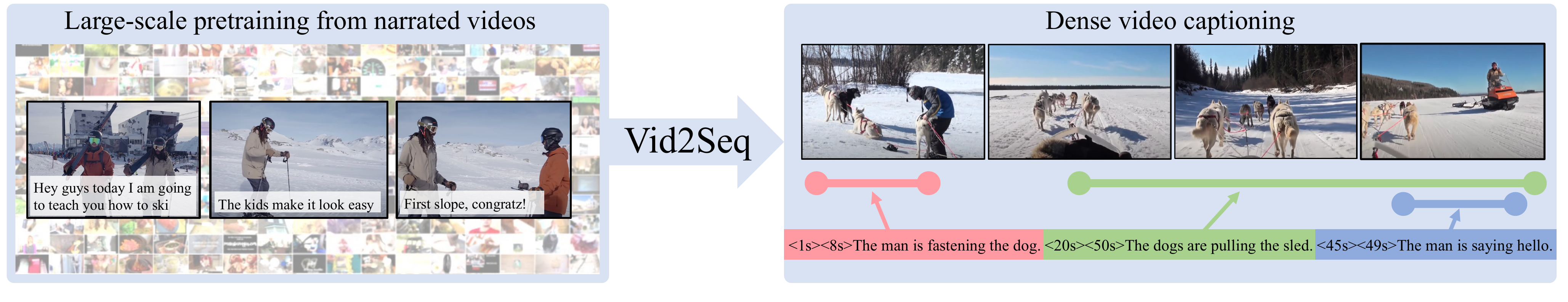}
\vspace{-0.8cm}
\captionof{figure}{\small 
\textbf{\model{}} is a visual language model that predicts dense event captions together with their temporal grounding in the video by generating a \emph{single} sequence of tokens (right).
This ability is enabled by large-scale pretraining on unlabeled narrated videos (left).
}
\vspace{-0.1cm}
\label{fig:teaser}
\end{center}
}]

\begin{abstract}
\vspace{-0.3cm}
In this work, we introduce \model{}, a multi-modal single-stage dense event captioning model pretrained on narrated videos which are readily-available at scale.
The \model{} architecture augments a language model with special time tokens, allowing it to seamlessly predict event boundaries and textual descriptions in the same output sequence.
Such a unified model requires large-scale training data, which is not available in current annotated datasets. 
We show that it is possible to leverage unlabeled narrated videos for dense video captioning, 
by reformulating sentence boundaries of transcribed speech as pseudo event boundaries, and using the transcribed speech sentences as pseudo event captions.
The resulting \model{} model pretrained on the YT-Temporal-1B dataset improves the state of the art on a variety of dense video captioning benchmarks including YouCook2, ViTT and ActivityNet Captions.
\model{} also generalizes well to the tasks of video paragraph captioning and video clip captioning, and to few-shot settings.
Our code is publicly available at~\cite{vid2seqwebpage}.
\end{abstract}

\renewcommand{\thefootnote}{\fnsymbol{footnote}}
\footnotetext[1]{This work was done when the first author was an intern at Google.}
\renewcommand*{\thefootnote}{\arabic{footnote}}

\vspace{-0.6cm}
\section{Introduction}\label{sec:intro}
\vspace{-0.1cm}
\input{intro.tex}

\section{Related Work}\label{sec:background}
\input{background.tex}

\section{Method}\label{sec:method}
\input{method.tex}

\vspace{-0.2cm}
\section{Experiments}\label{sec:experiments}
\input{results_arxiv.tex}

\vspace{-0.2cm}
\section{Conclusion}\label{sec:conclusion}
\input{conclusion.tex}

\input{ack.tex}

{\small
\bibliographystyle{ieee_fullname}
\bibliography{egbib}
}

\clearpage \newpage
\appendix

\section*{Appendix}
\input{appendix_arxiv.tex}

\end{document}

%% file: intro.tex
Dense video captioning requires the temporal localization and captioning of all events in an untrimmed video~\cite{krishna2017dense, wang2021end, zhou2018end}.
This differs from standard video captioning~\cite{lin2022swinbert, luo2020univilm, seo2022end}, where the goal is to produce a single caption for a given short video clip. 
Dense captioning is significantly more difficult, as it raises the additional complexity of localizing the events in minutes-long videos. 
However, it also benefits from long-range video information.
This task is potentially highly useful in applications such as large-scale video search and indexing, where the video content is not segmented into clips. 

Existing methods mostly resort to two-stage approaches~\cite{krishna2017dense, wang2018bidirectional, iashin2020better}, where events are first localized and then captioned.
To further enhance the inter-task interaction between event localization and captioning, some approaches have introduced models that jointly solve the two tasks~\cite{deng2021sketch, wang2021end, zhou2018end}.
However, often these approaches still require task-specific components such as event counters~\cite{wang2021end}. 
Furthermore, they exclusively train on manually annotated datasets of limited size~\cite{huang2020multimodal, krishna2017dense, youcook2}, which makes it difficult to effectively solve the task.
To address these issues, we take inspiration from recent sequence-to-sequence models pretrained on Web data which have been successful on a wide range of vision and language tasks~\cite{chen2021pix2seq, yang2021crossing, chen2022pali, alayrac2022flamingo, wang2021simvlm}.

First, we propose a video language model, called \model{}. 
We start from a language model trained on Web text~\cite{raffel2020exploring} and augment it with special \emph{time tokens} that represent timestamps in the video.
Given video frames and transcribed speech inputs, the resulting model jointly predicts all event captions and their corresponding temporal boundaries by generating a \emph{single} sequence of discrete tokens, as illustrated in Figure~\ref{fig:teaser} (right).
Such a model therefore has the potential to learn multi-modal dependencies between the different events in the video via attention~\cite{vaswani2017attention}.
However this requires large-scale training data, which is not available in current dense video captioning datasets~\cite{huang2020multimodal, krishna2017dense, youcook2}.
Moreover, collecting manual annotations of dense captions for videos is expensive and prohibitive at scale.

Hence we propose to pretrain \model{} by leveraging unlabeled narrated videos which are readily-available at scale.
To do this, we reformulate sentence boundaries of transcribed speech as pseudo event boundaries, and use the transcribed speech sentences as pseudo event captions.
We then pretrain \model{} with a generative objective, that requires predicting the transcribed speech given visual inputs, and a denoising objective, which masks spans of transcribed speech.
Note that transcribed speech may not describe the video content faithfully, and is often temporally misaligned with the visual stream~\cite{han2022temporal, ko2022video, miech20endtoend}. 
For instance, from the example in Figure~\ref{fig:teaser} (left), one can understand that the grey skier has descended a slope from the last speech sentence which is said \emph{after} he actually descended the slope.
Intuitively, \model{} is particularly suited for learning from such noisy supervision as it jointly models {\em all} narrations and the corresponding timestamps in the video.

We demonstrate the effectiveness of our pretrained model through extensive experiments.
We show 
the importance of pretraining on untrimmed narrated videos, 
the ability of \model{} to use both the visual and speech modalities, 
the importance of the pretraining objectives, 
the benefit of joint caption generation and localization, 
as well as the importance of the language model size and the scale of the pretraining dataset.
The pretrained \model{} model achieves state-of-the-art performance on various dense video captioning benchmarks~\cite{huang2020multimodal, krishna2017dense, youcook2}.
Our model also excels at generating paragraphs of text describing the video: without using ground-truth event proposals at inference time, our model outperforms all prior approaches including those that rely on such proposals~\cite{lei2020mart, zhou2019grounded, park2019adversarial}.
Moreover, \model{} generalizes well to the standard task of video clip captioning~\cite{chen2011collecting, xu16msrvtt}.
Finally, we introduce a new few-shot dense video captioning setting in which we finetune our pretrained model on a small fraction of the downstream training dataset and show benefits of \model{} in this setting.

In summary, we make the following contributions: 
\textit{(i)}~We introduce \model{} for dense video captioning. 
Given multi-modal inputs (transcribed speech and video), \model{} predicts a single sequence of discrete tokens that includes caption tokens interleaved with special \textit{time tokens} that represent event timestamps.
\textit{(ii)}~We show that transcribed speech and corresponding timestamps in unlabeled narrated videos can be effectively used as a source of weak supervision for dense video captioning.
\textit{(iii)}~Finally, our pretrained \model{} model improves the state of the art on three dense video captioning datasets (YouCook2, ViTT, ActivityNet Captions),
two video paragraph captioning benchmarks (YouCook2, ActivityNet Captions) 
and two video clip captioning datasets (MSR-VTT, MSVD), and also generalizes well to few-shot settings.

Our code implemented in Jax and based on the Scenic library~\cite{dehghani2021scenic} is publicly released at~\cite{vid2seqwebpage}.

%% file: background.tex
\noindent \textbf{Dense video captioning.} 
Dense video captioning lies at the intersection of event localization~\cite{heilbron2016fast, escorcia2016daps, gao2017turn, lin2018bsn, lin2019bmn, lin2020fast, shou2016temporal, zhao2017temporal} and event captioning~\cite{gao2017video, lin2022swinbert, pan2017video, wang2018video, wang2018reconstruction}.
The majority of existing methods for dense video captioning~\cite{krishna2017dense, iashin2020better, iashin2020multi, wang2018bidirectional, wang2020event} consist of a temporal localization stage followed by an event captioning stage.
To enrich inter-task interactions, recent works~\cite{chadha2020iperceive, chen2021towards, deng2021sketch, li2018jointly, mun2019streamlined, rahman2019watch, shen2017weakly, shi2019dense, wang2018bidirectional, wang2021end, zhou2018end} jointly train the captioning and localization modules.
In particular, Wang~\etal~\cite{wang2021end} propose to view dense video captioning as a set prediction task, and jointly perform event localization and captioning for each event in parallel.
In contrast, our model generates event boundaries and captions conditioned on the previously generated events.
Deng~\etal~\cite{deng2021sketch} propose to first generate a paragraph and then ground each sentence in the video.
We also generate all captions as a single output sequence, however our output already includes event timestamps.
Zhang~\etal~\cite{zhang2022unifying} propose to generate event boundaries sequentially, but separately perform event localization and single event captioning, and only use visual input. 
Most related to our work, Zhu~\etal~\cite{zhu2022end} also perform dense video captioning by generating a single output sequence. 
Their method, however, infers event locations directly from the timestamps of transcribed speech and, hence, can only detect events that closely follow the speech.
In contrast, our model generates event timestamps as special tokens and can produce dense captions for videos with limited speech, as we demonstrate on the ActivityNet Captions dataset.

\noindent \textbf{Video and language pretraining.} 
Following the success of image-text pretraining~\cite{singh2022flava, yu2022coca, hu2022scaling, chen2019uniter, dou2022empirical, dou2022coarse, gan2020large, huang2021seeing, jia2021scaling, mdetr, kim2021vilt, li2019unicodervl, li2020oscar, li2021align, li2022blip, li2022grounded, lu2019vilbert, lu202012, su2019vl, tan2019lxmert, tsimpoukelli2021multimodal,  desai2021redcaps, wang2021ufo, yu2020ernie, yuan2021florence, zhang2022glipv2, zhou2020unified}, recent works have explored video-text pretraining~\cite{wang2022object, akbari2021vatt, alayrac2022flamingo, bain2021frozen, fu2021violet, ge2022bridging, han2022temporal, ko2022video, lei2021less, li2020hero, li2021align2, miech19howto100m, miech20endtoend, nagrani2022learning, seo2020look, seo2022end, sun2019videobert, xue2022advancing, xu2021videoclip, wang2022all, yang2021just, Yang2022LearningTA, yang2022frozenbilm, yang2022tubedetr, zellers2021merlot, zellers2022merlot, xue2022advancing}.
These methods show strong improvements on various tasks such as text-video retrieval~\cite{bain2021frozen, miech20endtoend}, video question answering~\cite{yang2021just, zellers2021merlot} and video clip captioning~\cite{alayrac2022flamingo, seo2022end}.
While these works mostly learn global video representations to tackle video-level prediction tasks, we here focus on learning detailed representations to address a dense prediction task requiring reasoning over multiple events in untrimmed videos.
Several works have explored long-form video-text pretraining~\cite{sun2022long} and video-text pretraining for temporal localization tasks~\cite{cao2022locvtp, lei2021detecting, lin2022egocentric, wang2022contrastive, xu2022contrastive, yang2021taco}. 
However these works focus on video understanding tasks while our pretraining approach is tailored for a generative task that not only requires the model to reason over multiple events in the video, but also to describe them by natural language.

\begin{figure*}[t]
\centering
\includegraphics[clip, trim=2mm 4mm 3mm 2mm, width=1.0\linewidth]{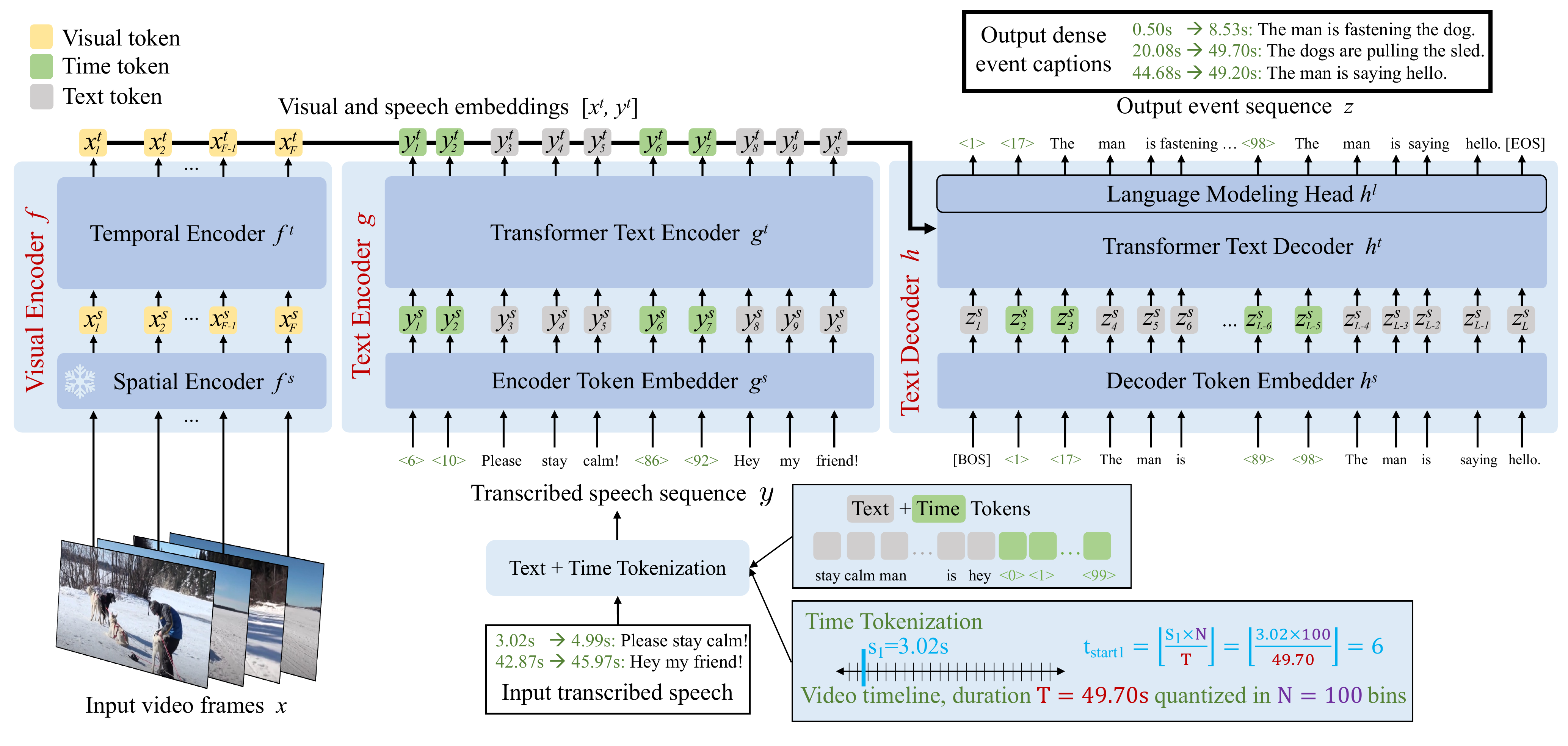}
\vspace{-0.5cm}
\caption{\small \textbf{\model{} model overview}.
We formulate dense event captioning as a sequence-to-sequence problem, using special \emph{time tokens} to allow the model to seamlessly understand and generate sequences of tokens containing both textual semantic information and temporal localization information grounding each text sentence in the video.
In detail, all input video frames $x$ and the transcribed speech sequence $y$ are first processed with a Visual Encoder $f$ (a frozen Spatial Encoder $f^s$ followed by a Temporal Encoder $f^t$) and a Text Encoder $g$ (a Token Embedder $g^s$ followed by a Transformer Encoder $g^t$), respectively.
Then the Text Decoder $h$ (composed of a Token Embedder $h^s$, a Transformer Encoder $h^t$ and a Language Modeling Head $h^l$) autoregressively generates the output event sequence $z$ by cross-attending to the visual and speech embeddings $x^{t}$ and $y^{t}$.
}
\vspace{-0.4cm}
\label{fig:overview}
\end{figure*}

A few works explore pretraining for dense video captioning.
Zhang~\etal~\cite{zhang2022unifying} pretrain on ActivityNet Captions to improve the downstream performance on the same dataset.
In contrast, we propose a pretraining method that does not rely on \emph{any} manual annotation, and show its benefits on multiple downstream datasets.
Huang~\etal~\cite{huang2020multimodal} explore pretraining on narrated instructional videos, but only consider event captioning using ground truth proposals as their model does not handle localization.
Finally, \cite{huang2020multimodal, zhu2022end} explore pretraining on a domain specific text-only dataset~\cite{koupaee2018wikihow}.
In contrast, we propose to pretrain on a generic video corpus~\cite{zellers2022merlot} and show benefits on various domains.

\noindent \textbf{Unifying tasks as language modeling.} 
Recent works~\cite{chen2021pix2seq, chen2022unified, chen2022obj2seq, chen2022pali, cho2021unifying, kolesnikov2022uvim, li2022label2label, wang2022unifying, yang2021crossing, zhu2022seqtr} have shown that it is possible 
to cast various computer vision problems as a language modeling task, addressing object detection~\cite{chen2021pix2seq}, grounded image captioning~\cite{yang2021crossing} or visual grounding~\cite{zhu2022seqtr}.
In this work we also cast visual localization as a language modeling task.
However, unlike prior work focused on image-level spatial localization, we address the different problem of event localization \textit{in time}, in untrimmed videos.

%% file: method.tex
The goal of dense video captioning is to temporally localize and describe with natural language {\em all} events in an untrimmed input video.
Therefore a key challenge is to effectively model the relationships between the different events in the video, as for example, it is easier to predict that the dogs are pulling the sled if we know that the man has just fastened a dog (see Figure~\ref{fig:teaser} (right)).
Furthermore, due to the dense nature of the task, there can be many events in a long video and the requirement is to output a natural language caption for each event. Hence, another key challenge is that the manual collection of annotations for this task is particularly expensive.
To tackle these challenges, we first develop a unified multi-modal model that jointly predicts event boundaries and captions as a single sequence of tokens, as explained in Section~\ref{sec:model} and Figure~\ref{fig:overview}.
Second, we design a pretraining strategy that effectively leverages cross-modal supervision in the form of transcribed speech from unlabeled narrated videos by reformulating sentence boundaries as pseudo event boundaries, as presented in Section~\ref{sec:training} and Figure~\ref{fig:pretraining}.

\vspace{-0.2cm}
\subsection{Model}\label{sec:model}
We wish to design a model for dense video captioning that can capture relationships between events using visual and (transcribed) speech cues
in order to effectively localize and describe these events in untrimmed minutes-long videos.
To tackle this challenge, we cast dense video captioning as a sequence-to-sequence problem where the input and output sequences contain both the semantic information about the event in the form of natural language descriptions and the temporal localization of the events in the form of temporal timestamps. 
In addition, to best leverage both the visual and the language signal, we develop an appropriate multi-modal encoder-decoder architecture.
As illustrated in Figure~\ref{fig:overview}, our architecture takes as input video frames $x=\{x_i\}_{i=1}^{F}$ together with the transcribed speech sequence $y=\{y_j\}_{j=1}^{S}$. The output of our model is an event sequence $z=\{z_k\}_{k=1}^{L}$, where each event contains both its textual description and timestamps corresponding to the temporal event locations in the video.
Below we explain the structure of the transcribed speech and event sequences constructed for our model as well as details of our model architecture.

\vspace{-0.1cm}
\paragraph{Sequence construction.}\label{sec:sequence}
To model inter-event relationships in dense event captioning annotations (or the readily-available transcribed narration, see Section~\ref{sec:training}), we cast dense video captioning as predicting a single output sequence of tokens $z$.
This output event sequence is constructed by leveraging a text tokenizer augmented with special \emph{time tokens}.
Furthermore, we enable our architecture to jointly reason about the semantic and temporal information provided in the transcript of the input narration by constructing the input transcript sequence $y$ in a similar manner as the event sequence $z$.
Details are given next.

\noindent \textbf{\textit{Time tokenization.}}
We start from a text tokenizer with a vocabulary size $V$, and augment it with $N$ additional time tokens, resulting in a tokenizer with $V+N$ tokens. 
The time tokens represent relative timestamps in a video, as we quantize a video of duration $T$ into $N$ equally-spaced timestamps.
In detail, we use the SentencePiece tokenizer~\cite{kudo2018sentencepiece} with vocabulary size $V=32,128$ and $N=100$.

\noindent \textbf{\textit{Event sequence.}} 
Our introduced tokenizer enables us to construct sequences that contain both video timestamps and text video descriptions.
We next explain how we construct the output event sequence $z$.
Note that videos have a variable number of events in standard dense video captioning datasets~\cite{huang2020multimodal, krishna2017dense, youcook2}.
Each event $k$ is characterized by a text segment, a start time and an end time.
We first construct for each event $k$ a sequence by concatenating its start time token $t_{start_k}$, its end time token $t_{end_k}$ and its text tokens $[z_{k_1}, ..., z_{k_{l_k}}]$.
Then we order all these sequences in increasing order of their start times and concatenate them.
In practice, each text segment ends with a dot symbol indicating the separation between different events.
Finally, the event sequence is obtained by prepending and appending a BOS and an EOS tokens to indicate the start and the end of sequence, respectively, \ie $z=[BOS, t_{start_1}, t_{end_1}, z_{1_1}, ..., z_{1_{l_1}}, t_{start_2}, ..., EOS]$.

\footnotetext[1]{https://cloud.google.com/speech-to-text/docs/automatic-punctuation.}

\noindent \textbf{\textit{Transcribed speech sequence.}}
To enable the model to use both the transcribed speech and its corresponding timestamps, we convert the speech transcript into a speech sequence $y$ similarly as the input training dense event captions $z$.
This is done by segmenting the raw speech transcript into sentences with the Google Cloud API$^1$, and using each transcribed speech sentence with its corresponding timestamps analogously as an event in the previously explained process.

\paragraph{Architecture.}\label{sec:architecture}
We wish to design an architecture that can effectively model relationships between different events in untrimmed minutes-long videos.
To tackle this challenge, we propose a multi-modal encoder-decoder architecture, illustrated in Figure~\ref{fig:overview}, that 
gradually refines and outputs the event sequence described above. 
In detail, given an untrimmed minutes-long video, the visual encoder $f$ embeds its frames while the text encoder $g$ embeds transcribed speech and the corresponding timestamps. 
Then a text decoder $h$ predicts event boundaries and text captions using the visual and transcribed speech embeddings. 
The individual modules are described next. 

\noindent \textit{\textbf{Visual encoder.}}
The visual encoder operates on a sequence of $F$ frames $x \in \mathbb{R}^{F \times H \times W \times C}$ where $H$, $W$ and $C$ are the height, width and the number of channels of each frame.
A visual backbone $f^s$ first encodes each frame separately and outputs frame embeddings $x^s = f^s(x) \in \mathbb{R}^{F \times d}$, where $d$ is the embedding dimension.
Then a transformer encoder~\cite{vaswani2017attention} $f^t$ models temporal interactions between the different frames, and outputs $F$ contextualized visual embeddings $x^t = f^t(x^s + x^p) \in \mathbb{R}^{F \times d}$, where $x^p \in \mathbb{R}^{F \times d}$ are learnt temporal positional embeddings, which communicate time information from visual inputs to the model.
\noindent In detail, the visual backbone is CLIP ViT-L/14~\cite{dosovitskiy2021an, radford2021learning} at resolution $224\times224$ pixels, pretrained to map images to text descriptions with a contrastive loss on Web-scraped image-text pairs. 
We keep the backbone frozen for efficiency.

\noindent \textit{\textbf{Text encoder.}}
The text encoder operates on a transcribed speech sequence of $S$ tokens $y \in  \{1, ..., V+N\}^S$, where $V$ is the text vocabulary size, $N$ is the size of the vocabulary of time tokens and $S$ is the number of tokens in the transcribed speech sequence.
Note that the transcribed speech sequence includes time tokens to input the temporal  information from the transcribed speech into the model.
An embedding layer $g^s \in \mathbb{R}^{(V+N) \times d}$ embeds each token independently and outputs semantic embeddings $y^s=g^s(y) \in \mathbb{R}^{S \times d}$.
Then a transformer encoder $g^t$ computes interactions in the transcribed speech sequence and outputs $S$ contextualized speech embeddings $y^t=g^t(y^s) \in \mathbb{R}^{S \times d}$.

\noindent \textit{\textbf{Text decoder.}}
The text decoder generates the event sequence $z$ by using the encoder embeddings, which are obtained by concatenating the visual and speech embeddings $x^t$ and $y^t$.
The text decoder is based on a causal transformer decoder $h^t$ that cross-attends to the encoder outputs, and at each autoregressive step $k$, self-attends to the previously generated tokens $\hat{z}^{t}_{<k}$ to output a contextualized representation $z^t_k=h^t(h^s(\hat{z}^{t}_{<k}), x^t, y^t) \in \mathbb{R}^{d}$ where $h^s \in \mathbb{R}^{(V+N) \times d}$ is the decoder token embedding layer.
Then a language modeling head $h^{l} \in \mathbb{R}^{d \times (V+N)}$ predicts a probability distribution over the joint vocabulary of text and time tokens in order to predict the next token in the event sequence, \ie $z^{l}_k = h^{l}(z^t_k) \in \mathbb{R}^{V + N}$.

\noindent \textit{\textbf{Text initialization.}}
We initialize the text encoder and the text decoder with T5-Base~\cite{raffel2020exploring} which has been pretrained on Web text corpora with a denoising loss.
Therefore their implementation and parameters also closely follow T5-Base, \eg they use relative positional embeddings and share their token embedding layer $g^s = h^s \in \mathbb{R}^{(V+N) \times d}$.

\subsection{Training}\label{sec:training}
\begin{figure}[t]
\centering
\includegraphics[width=1.\linewidth]{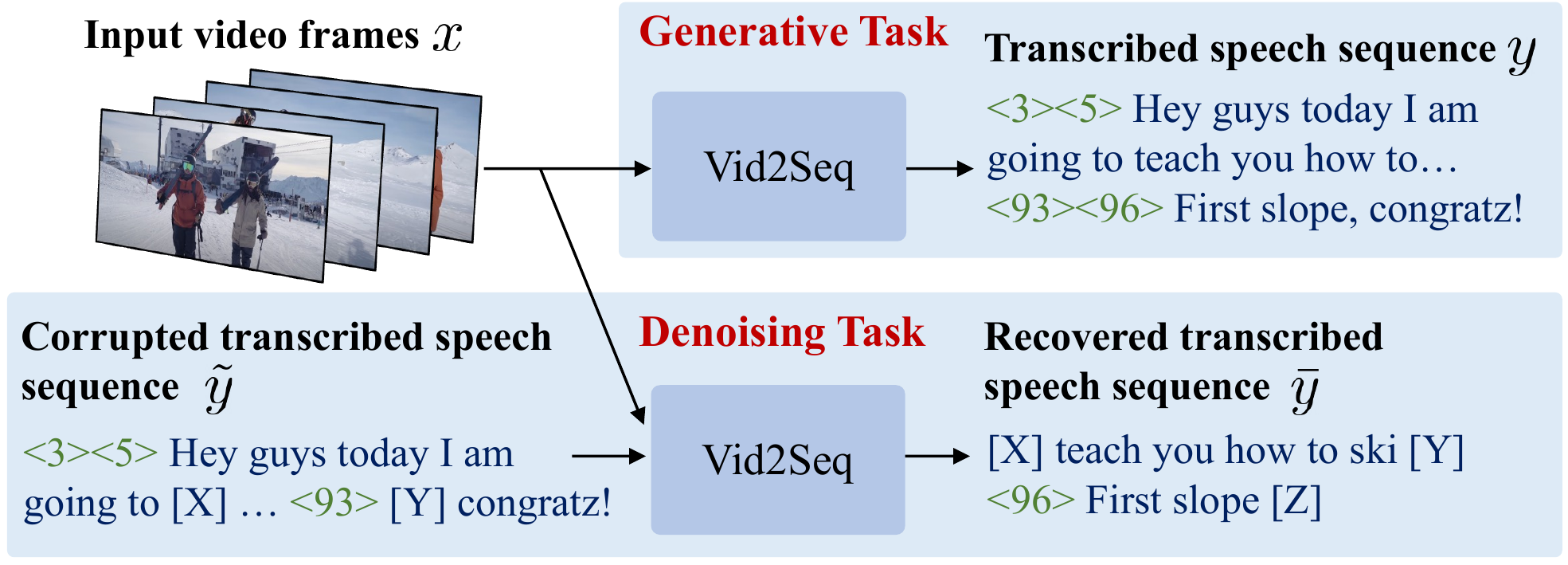}
\vspace{-0.7cm}
\caption{\small \textbf{Pretraining tasks}. 
To train \model{} on unlabeled narrated videos, we design two pretraining objectives.
\textbf{Top}: generative objective, 
given visual inputs $x$ only, the task is to generate the transcribed speech sequence $y$.
\textbf{Bottom}: denoising objective,
given visual inputs $x$ and the corrupted 
speech sequence $\tilde{y}$, the task is to generate the sequence of recovered 
speech segments $\Bar{y}$.}
\label{fig:pretraining}
\vspace{-0.5cm}
\end{figure}

In this Section, we describe how we leverage a large amount of unlabeled narrated videos to train the previously described dense event captioning model.
We first present the pretraining method used to effectively train \model{} using cross-modal supervision in readily-available narrated videos in Section~\ref{sec:pretraining} and Figure~\ref{fig:pretraining}.
Then we explain how we finetune our architecture for various downstream tasks including dense event captioning in Section~\ref{sec:downstream}.

\vspace{-0.3cm}
\subsubsection{Pretraining on untrimmed narrated videos}\label{sec:pretraining}
We wish to leverage narrated videos for pretraining as they are easily available at scale~\cite{miech19howto100m, zellers2022merlot}.
However these videos do not contain dense event captioning annotations.
Therefore we use as supervisory signal the transcribed speech sentences and their corresponding timestamps.
As speech transcripts are not always visually grounded and often temporally misaligned~\cite{han2022temporal, ko2022video, miech20endtoend}, we note that they only provide \emph{weak} supervision.
Furthermore, speech transcripts drastically differ from dense event captioning annotations.
For instance, in the YT-Temporal-1B dataset~\cite{zellers2022merlot}, a video contains 120 speech sentences on average which is an order of magnitude more than the number of events in standard dense video captioning datasets~\cite{youcook2, huang2020multimodal, krishna2017dense}.
Our \model{} model is particularly suitable for using such weak supervision as it constructs the speech sequence similarly as a manually annotated event sequence, 
and jointly contextualizes the speech boundaries and semantic information on the level of potentially minutes-long videos (see Section~\ref{sec:model}) rather than at a shorter clip-level,
enabling our model to learn long-term relationships between the different speech segments: in experiments we show that pretraining on entire minutes-long videos is highly beneficial.

We next describe the two proposed training objectives, which are both based on a maximum likelihood objective.
Formally, given visual inputs $x$, encoder text sequence $y$ and a decoder target text sequence $z$, both objectives are based on minimizing the following loss:
\setlength{\abovedisplayskip}{0.pt} 
\setlength{\belowdisplayskip}{0.pt}
\begin{equation}
{\mathcal{L}_\theta(x, y, z)} = - \frac{1}{\sum_{k=1}^{L-1}{w_k}}\sum_{k=1}^{L-1}{w_k \log\,p_\theta(z_{k+1} | x, y, z_{1:k})},
\label{eq:loss}
\end{equation}
where $L$ is the length of the decoder target sequence, $w_k$ is the weight for k-th token in the sequence, which we set to $w_k=1$ $\forall k$ in practice, $\theta$ denotes the trainable parameters in the model and $p_\theta$ is the output probability distribution over the vocabulary of text and time tokens.

\noindent\textbf{Generative objective.}
This objective uses the transcribed speech as a (pseudo-)supervisory signal to teach the decoder to predict a sequence of events given visual inputs.
Given video frames $x$, which are fed to the encoder, the decoder has to predict the transcribed speech sequence $y$ (see Figure~\ref{fig:pretraining}), which serves as a proxy dense event captioning annotation.
Note that no text input is given to the encoder for this task as using transcribed speech both as input and target would lead the model to learn text-only shortcuts.

\noindent \textbf{Denoising objective.}
As no text input is given to the encoder for the generative proxy task, the generative objective only trains the visual encoder and the text decoder, but not the text encoder.
However when our model is used for dense video captioning, the text encoder has a significant importance as it encodes speech transcripts.
Hence we introduce a denoising objective that aims at jointly aligning the visual encoder, the text encoder and the text decoder.
Inspired by T5~\cite{raffel2020exploring} in the text domain, we randomly mask spans of (text and time) tokens in the transcribed speech sequence with a probability $P$ and an average span length $M$.
The encoder input is composed of the video frames $x$ together with the corrupted speech sequence $\tilde{y}$, which contains sentinel tokens that uniquely identify the masked spans.
The decoder then has to predict a sequence $\Bar{y}$ constructed with the corresponding masked spans for each sentinel token, based on visual inputs $x$ and speech context $\tilde{y}$ (see Figure~\ref{fig:pretraining}).

\vspace{-0.3cm}
\subsubsection{Downstream task adaptation}\label{sec:downstream}
Our architecture and task formulation enables us to tackle dense video captioning with a generic language modeling training objective and inference procedure.
Note that as a by-product of our generic architecture, our model can also be used to generate paragraphs about entire videos by simply removing the time tokens from the output sequence, 
and can also be easily adapted to video clip captioning with the same finetuning and inference recipe.

\noindent \textbf{Finetuning.}
To finetune our model for dense video captioning, we use a maximum likelihood objective based on the event sequence (see Equation~\ref{eq:loss}).
Given video frames $x$ and speech transcripts $y$, the decoder has to predict the event sequence $z$.

\noindent \textbf{Inference.}
The text decoder autoregressively generates the event sequence by sampling from the model likelihood.
In practice, we use beam search as we find that it improves the captioning quality compared with argmax sampling or nucleus sampling.
Finally, the event sequence is converted into a set of event predictions by simply reversing the sequence construction process.

%% file: results_arxiv.tex
\vspace{-0.1cm}

This section demonstrates the effectiveness of our pretrained \model{} model and compares our method to the state of the art.
We first outline our experimental setup in Section~\ref{sec:setup}.
We then present ablation studies in Section~\ref{sec:ablation}.
The comparison to the state of the art in dense video captioning, video paragraph captioning and video clip captioning is presented in Section~\ref{sec:sota}. 
Next, we present results in a new few-shot dense video captioning setting in Section~\ref{sec:fewshot}. 
Finally, we show qualitative results in Section~\ref{sec:qualitative}. 

\subsection{Experimental setup}\label{sec:setup}
\vspace{-0.1cm}

\noindent \textbf{Datasets.} 
For pretraining, following prior work showing the benefits of pretraining on a diverse and large dataset~\cite{zellers2021merlot}, we use the \textbf{YT-Temporal-1B} dataset~\cite{zellers2022merlot}, which includes 18 million narrated videos collected from YouTube.
We evaluate \model{} on three downstream dense video captioning datasets: YouCook2~\cite{youcook2}, ViTT~\cite{huang2020multimodal} and ActivityNet Captions~\cite{krishna2017dense}.
\textbf{YouCook2} has 2K untrimmed videos of cooking procedures.
On average, each video lasts 320s and is annotated with 7.7 temporally-localized sentences.
\textbf{ViTT} consists of 8K untrimmed instructional videos.
On average, each video lasts 250s and is annotated with 7.1 temporally-localized short tags.
\textbf{ActivityNet Captions} contains 20k untrimmed videos of various human activities.
On average, each video lasts 120s and is annotated with 3.7 temporally-localized sentences. 
For video clip captioning, we use two standard benchmarks, \textbf{MSR-VTT}~\cite{xu16msrvtt} and \textbf{MSVD}~\cite{chen2011collecting}.
For all datasets, we follow the standard splits for training, validation and testing.
Note that we only use videos available on YouTube at the time of the work, resulting in 10 to 20\% less videos than in the original datasets.

\noindent \textbf{Implementation details.} 
We extract video frames at 1FPS, and subsample or pad the sequence of frames to $F$ frames where we set $F=100$.
The text encoder and decoder sequence are truncated or padded to $L=S=1000$ tokens.
Our model has 314M trainable parameters.
We use the Adam optimizer~\cite{kingma15adam}.
We pretrain our model for 200,000 iterations with a batch size of 512 videos split on 64 TPU v4 chips, which lasts a day.
We sum both pretraining objectives with equal weighting to get our final pretraining loss.
More details are included in Appendix Section~\ref{sec:adddetails}.

\begin{table}[t]
\centering
\vspace{-0pt}
\begin{center}
\setlength\tabcolsep{6pt}
\resizebox{1.\linewidth}{!}{
\begin{tabular}{ccc|ccc|ccc}
\toprule
& \multicolumn{2}{c|}{Pretraining input} & \multicolumn{3}{c|}{YouCook2} & \multicolumn{3}{c}{ActivityNet} \\ & Untrimmed & Time tokens & \small{S} & \small{C} & \small{F1} & \small{S} & \small{C} & \small{F1} \\
\midrule
1. & \multicolumn{2}{c|}{\textit{No pretraining}}
& 4.0 & 18.0 & 18.1 
& 5.4 & 18.8 & 49.2 \\ 
2. & \xmark & \xmark
& 5.5 & 27.8 & 20.5
& 5.5 & 26.5 & 52.1 \\ 
3. & \cmark & \xmark
& 6.7 & 35.0 & 23.3 
& 5.6 & 27.4 & 52.2 \\ 
4. & \cmark & \cmark
& \textbf{7.9} & \textbf{47.1} & \textbf{27.3} 
& \textbf{5.8} & \textbf{30.1} & \textbf{52.4} \\ 
\bottomrule
\end{tabular}}
\vspace{-0.3cm}
\caption{\small \textbf{Ablation showing the impact of using untrimmed videos and adding time tokens during pretraining.} When we use untrimmed video-speech inputs, time information from transcribed speech sentence boundaries is integrated via time tokens.}
\vspace{-0.9cm}
\label{table:pretraining}
\end{center}
\end{table}

\noindent \textbf{Evaluation metrics.} 
For captioning, we use
CIDEr~\cite{vedantam2015cider} (C) and METEOR~\cite{banerjee2005meteor} (M).
For dense video captioning, we follow the commonly used evaluation tool~\cite{krishna2017dense} which calculates matched pairs between generated events and the ground truth across IoU thresholds of \{0.3, 0.5, 0.7, 0.9\}, and compute captioning metrics over the matched pairs.
However, these metrics do not take into account the story of the video.
Therefore we also use SODA\_c~\cite{fujita2020soda} (S) for an overall dense video captioning evaluation.
To further isolate the evaluation of event localization, we report the average precision and average recall across IoU thresholds of \{0.3, 0.5, 0.7, 0.9\} and their harmonic mean, the F1 Score.

\subsection{Ablation studies}\label{sec:ablation}
\vspace{-0.1cm}

The default \model{} model predicts both text and time tokens, uses both visual frames and transcribed speech as input, builds on the T5-Base language model, and is pretrained on untrimmed videos from YT-Temporal-1B with both the generative and denoising losses.
Below we ablate the importance of each of these factors on the downstream dense video captioning performance by reporting results on YouCook2 and ActivityNet Captions validation sets. 

\noindent \textbf{Pretraining on untrimmed narrated videos by exploiting transcribed speech sentence boundaries.} 
In Table~\ref{table:pretraining}, we evaluate the effectiveness of our pretraining task formulation that uses untrimmed videos and integrates sentence boundaries of transcribed speech via time tokens.
In contrast, most video clip captioning pretraining methods~\cite{huang2020multimodal, luo2020univilm, seo2022end} use short, trimmed, video-speech segments for pretraining.
We adapt this strategy in our model and find that it indeed yields significant performance improvements over the baseline that uses no video-text pretraining (row 2 vs row 1).
However, larger improvements are obtained by using untrimmed video-speech inputs (row 3 vs row 2).
Moreover, using time tokens to integrate time information from transcribed speech drastically improves performance (row 4 vs row 3).
This shows the benefits of exploiting sentence boundaries of transcribed speech via time tokens and of using untrimmed videos during pretraining.
In Appendix Section~\ref{sec:ablation2}, we show additional ablations to quantify how the performance improves by pretraining on longer narrated videos that contain more speech sentences.

\begin{table}[t]
\centering
\vspace{-0pt}
\begin{center}
\setlength\tabcolsep{3pt}
\resizebox{1.\linewidth}{!}{
\begin{tabular}{ccc|cc|ccc|ccc}
\toprule
& \multicolumn{2}{c|}{Finetuning Input} & \multicolumn{2}{c|}{Pretraining loss} & \multicolumn{3}{c|}{YouCook2} & \multicolumn{3}{c}{ActivityNet} \\ 
& Visual & Speech & Generative & Denoising & \small{S} & \small{C} & \small{F1} & \small{S} & \small{C} & \small{F1} \\
\midrule
1. & \cmark & \xmark & \multicolumn{2}{c|}{\textit{No pretraining}}
& 3.0 & 15.6 & 15.4 
& 5.4 & 14.2 & 46.5 \\
2. & \cmark & \cmark & \multicolumn{2}{c|}{\textit{No pretraining}}
& 4.0 & 18.0 & 18.1 
& 5.4 & 18.8 & 49.2 \\ 
3. & \cmark & \xmark & \cmark & \xmark
& 5.7 & 25.3 & 23.5 
& \textbf{5.9} & \textbf{30.2} & 51.8 \\
4. & \cmark & \cmark & \cmark & \xmark
& 2.5 & 10.3 & 15.9 
& 4.8 & 17.0 & 48.8 \\
5. & \cmark & \cmark & \cmark & \cmark
& \textbf{7.9} & \textbf{47.1} & \textbf{27.3} 
& 5.8 & 30.1 & \textbf{52.4} \\ 
\bottomrule
\end{tabular}}
\vspace{-0.3cm}
\caption{\small \textbf{Effect of input modalities and pretraining losses.}}
\vspace{-0.7cm}
\label{table:modalities}
\end{center}
\end{table}

\begin{table}[t]
\centering
\vspace{-0pt}
\begin{center}
\setlength\tabcolsep{3pt}
\resizebox{1.\linewidth}{!}{
\begin{tabular}{ccc|ccc|ccc}
\toprule
& \multirow{2}{*}{Captioning} & \multirow{2}{*}{Pretraining}
& \multicolumn{3}{c|}{YouCook2}
& \multicolumn{3}{c}{ActivityNet} \\
& & & \small{Recall} & \small{Precision} &
\small{F1} & \small{Recall} & \small{Precision} &
\small{F1} \\
\midrule
1. & \xmark & \xmark
& 17.8 & 19.4 & 17.7 & 47.3 & 57.9 & 52.0 \\ 
2. & \cmark & \xmark
& 17.2 & 20.6 & 18.1 & 42.5 & \textbf{64.1} & 49.2 \\ 
3. & \xmark & \cmark
& 25.7 & 21.4 & 22.8 & 52.5 & 53.0 & 51.1 \\ 
4. & \cmark & \cmark
& \textbf{27.9} & \textbf{27.8} & \textbf{27.3} & \textbf{52.7} & 53.9 & \textbf{52.4} \\ 
\bottomrule
\end{tabular}}
\vspace{-0.3cm}
\caption{\small \textbf{Effect of joint captioning and localization on the localization performance.}
The variant that does not caption corresponds to a localization-only variant that only predicts time tokens.}
\vspace{-0.7cm}
\label{table:inter}
\end{center}
\end{table}

\begin{table}[t]
\centering
\vspace{-0pt}
\begin{center}
\setlength\tabcolsep{4pt}
\resizebox{1.\linewidth}{!}{
\begin{tabular}{cc|cc|ccc|ccc}
\toprule
& \multirow{2}{*}{\makecell{\small{Language} \\ \small{Model}}} & \multicolumn{2}{c|}{Pretraining} & \multicolumn{3}{c|}{YouCook2} & \multicolumn{3}{c}{ActivityNet} \\
& & \# Videos & Dataset & \small{S} & \small{C} & \small{F1} 
& \small{S} & \small{C} & \small{F1} \\
\midrule
1. & T5-Small & 15M & YTT
& 6.1 & 31.1 & 24.3 
& 5.5 & 26.5 & 52.2 \\
2. & T5-Base & $\emptyset{}$ & $\emptyset{}$
& 4.0 & 18.0 & 18.1
& 5.4 & 18.8 & 49.2 \\ 
3. & T5-Base & 15K & YTT
& 6.3 & 35.0 & 24.4 
& 5.1 & 24.4 & 49.9 \\ 
4. & T5-Base & 150K & YTT
& 7.3 & 40.1 & 26.7
& 5.4 & 27.2 & 51.3 \\ 
5. & T5-Base & 1M5 & YTT
& 7.8 & 45.5 & 26.8
& 5.6 & 28.7 & 52.2 \\ 
6. & T5-Base & 1M & HTM
& \textbf{8.3} & \textbf{48.3} & 26.6 
& \textbf{5.8} & 28.8 & \textbf{53.1} \\ 
7. & T5-Base & 15M & YTT
& 7.9 & 47.1 & \textbf{27.3} 
& \textbf{5.8} & \textbf{30.1} & 52.4 \\ 
\bottomrule
\end{tabular}}
\vspace{-0.3cm}
\caption{\small\textbf{Effect of language model size and pretraining data.} HTM: HowTo100M~\cite{miech19howto100m}, YTT: YT-Temporal-1B~\cite{zellers2022merlot}.} 
\vspace{-0.9cm}
\label{table:scale}
\end{center}
\end{table}

\noindent \textbf{Input modalities and pretraining objectives.} 
In Table~\ref{table:modalities}, we analyze the importance of input modalities and pretraining tasks on the downstream dense video captioning performance.
The model with visual inputs only (no transcribed speech as input) benefits significantly from pretraining with the generative objective (row 3 vs row 1).
This shows the effectiveness of using the transcribed speech as a proxy annotation for dense video captioning pretraining.
However, this model is pretrained with visual inputs only and its performance largely drops when it is finetuned with both visual and transcribed speech inputs (row 4 vs row 3).
With both modalities, adding the denoising loss strongly benefits our model (row 5 vs rows 4 and 2).
We conclude that the denoising objective benefits multi-modal reasoning.

\noindent \textbf{Effect of captioning on localization.}
In Table~\ref{table:inter}, we compare the event localization performance of our model with a localization-only variant that only predicts event boundaries.
We find that the model that jointly predicts event boundaries and captions localizes better and benefits more from pretraining than the localization-only baseline (row 4 vs row 3), which demonstrates the importance of contextualizing the noisy timestamps of the transcribed speech with the speech semantic content during pretraining.

\noindent \textbf{Model size and pretraining data.}
In Table~\ref{table:scale}, we show that the language model size has a great importance on the performance, as the model with T5-Base outperforms its variant with T5-Small (row 7 vs row 1).
We also evaluate the importance of the size of the pretraining dataset of narrated videos by constructing subsets such that larger subsets include the smaller ones.
We find that scaling up the size of the pretraining dataset is beneficial, and that our pretraining method yields important benefits when only using 150K narrated videos for pretraining (row 4).
We further show that our pretraining method generalizes well to the HowTo100M dataset~\cite{miech19howto100m}.
The model pretrained on HowTo100M (row 6) actually achieves best results on YouCook2, as these datasets are from a similar domain.
Finally, we ablate the importance of the size and pretraining of the visual backbone in Appendix Section~\ref{sec:ablation2}.

\begin{table}[t]
\begin{center}
\setlength\tabcolsep{4pt}
\resizebox{1.\linewidth}{!}{
\begin{tabular}{l|c|ccc|ccc|ccc}
\toprule
\multirow{2}{*}{Method} & \multirow{2}{*}{Backbone}
& \multicolumn{3}{c|}{YouCook2 (val)}
& \multicolumn{3}{c|}{ViTT (test)}
& \multicolumn{3}{c}{ActivityNet (val)} \\
& & \small{S} & \small{C} & \small{M}
& \small{S} & \small{C} & \small{M} 
& \small{S} & \small{C} & \small{M} \\
\midrule
MT~\cite{zhou2018end} & TSN
& --- & 6.1 & 3.2
& --- & --- & ---
& --- & 9.3 & 5.0 \\
ECHR~\cite{wang2020event} & C3D
& --- & --- & 3.8
& --- & --- & ---
& 3.2 & 14.7 & 7.2 \\
PDVC~\cite{wang2021end} & TSN
& 4.4 & 22.7 & 4.7
& --- & --- & ---
& 5.4 & 29.0 & 8.0 \\
PDVC~\cite{wang2021end}$^\dag$ & CLIP
& 4.9 & 28.9 & 5.7
& --- & --- & ---
& \textbf{6.0} & 29.3 & 7.6 \\
UEDVC~\cite{zhang2022unifying} & TSN
& --- & --- & ---
& --- & --- & ---
& 5.5 & --- & --- \\
E2ESG~\cite{zhu2022end} & C3D
& --- & 25.0* & 3.5
& --- & 25.0 & 8.1
& --- & --- & ---- \\
\model{} (Ours) & CLIP
& \textbf{7.9} & \textbf{47.1} & \textbf{9.3}
& \textbf{13.5} & \textbf{43.5} & \textbf{8.5}
& 5.8 & \textbf{30.1} & \textbf{8.5} \\
\bottomrule
\end{tabular}
}
\vspace{-0.3cm}
\caption{\small Comparison to the state of the art for dense video captioning. * Results provided by the authors. $^\dag$ Results of our experiments using the official codebase.}
\label{table:sotac}
\vspace{-0.7cm}
\end{center}
\end{table}

\begin{table}[t]
\begin{center}
\setlength\tabcolsep{5pt}
\resizebox{1.\linewidth}{!}{
\begin{tabular}{l|c|cc|cc|cc}
\toprule
\multirow{2}{*}{Method} & \multirow{2}{*}{Backbone}
& \multicolumn{2}{c|}{YouCook2 (val)} 
& \multicolumn{2}{c|}{ViTT (test)}
& \multicolumn{2}{c}{ActivityNet (val)} \\
& & \small{Recall} & \small{Precision}
& \small{Recall} & \small{Precision}
& \small{Recall} & \small{Precision} \\
\midrule
PDVC~\cite{wang2021end} & TSN
& --- & ---
& --- & ---
& 55.4 & 58.1 \\
PDVC~\cite{wang2021end}$^\dag$ & CLIP
& --- & ---
& --- & ---
& 53.2 & 54.7 \\
UEDVC~\cite{zhang2022unifying} & TSN
& --- & ---
& --- & ---
& \textbf{59.0} & \textbf{60.3} \\
E2ESG~\cite{zhu2022end} & C3D
& 20.7* & 20.6*
& 32.2* & 32.1*
& --- & --- \\
\model{} (Ours) & CLIP
& \textbf{27.9} & \textbf{27.8}
& \textbf{42.6} & \textbf{46.2}
& 52.7 & 53.9 \\
\bottomrule
\end{tabular}
}
\vspace{-0.3cm}
\caption{\small Comparison to the state of the art for event localization. * Results provided by the authors. $^\dag$ Results of our experiments using the official codebase.}  
\label{table:sotal}
\end{center}
\vspace{-0.9cm}
\end{table}

\subsection{Comparison to the state of the art}\label{sec:sota}
\noindent \textbf{Dense video captioning.} 
In Table~\ref{table:sotac}, we compare our approach to state-of-the-art dense video captioning methods using cross-entropy training~\footnote{We do not include methods directly optimizing the test metric~\cite{deng2021sketch, mun2019streamlined}.} on the YouCook2, ViTT and ActivityNet Captions datasets.
\model{} sets new state of the art on all three datasets.
In particular, our method improves the CIDEr metric by 18.2 and 0.8 points on YouCook2 and ActivityNet Captions over PDVC.
Our method also outperforms E2ESG~\cite{zhu2022end} which uses in-domain text-only pretraining on Wikihow.
These results demonstrate the strong dense event captioning ability of our pretrained \model{} model.

\begin{figure*}[t]
\centering
\includegraphics[clip, trim=0mm 9cm 0mm 0mm, width=1\linewidth]{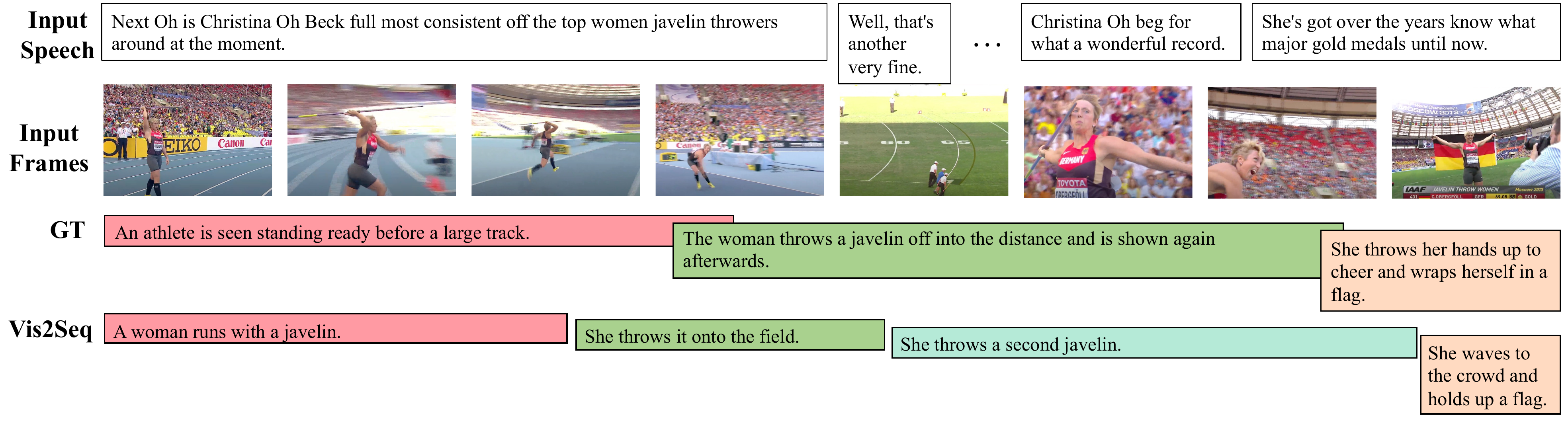}
\vspace{-0.9cm}
\caption{\small Example of dense event captioning predictions of \model{} on ActivityNet Captions validation set, compared with ground-truth.
}
\vspace{-0.6cm}
\label{fig:qualitative}
\end{figure*}

\noindent \textbf{Event localization.} 
In Table~\ref{table:sotal}, we evaluate the event localization performance of our dense video captioning model in comparison with prior work.
On both YouCook2 and ViTT, \model{} outperforms prior work~\cite{zhu2022end} tackling dense video captioning as a single sequence generation task.
However, our model underperforms compared to PDVC~\cite{wang2021end} and UEDVC~\cite{wang2021end} on ActivityNet Captions.
We emphasize that our approach integrates less prior knowledge about temporal localization than both these approaches, which include task specific components such as event counters~\cite{wang2021end} or separately train a model for the localization subtask~\cite{zhang2022unifying}.

\begin{table}[t]
\begin{center}
\setlength\tabcolsep{1pt}
\resizebox{1\linewidth}{!}{
\begin{tabular}{l|c|cc|cc}
\toprule
\multirow{2}{*}{Method} & \multirow{2}{*}{Backbone} 
& \multicolumn{2}{c|}{YouCook2 (val)}
& \multicolumn{2}{c}{ActivityNet (val-ae)} \\
& & \small{C} & \small{M} & \small{C} & \small{M} \\
\midrule
\textit{With GT Proposals} & & & \\
VTransformer~\cite{zhou2018end} & V (ResNet-200) + F %
& 32.3 & 15.7 & 22.2 & 15.6 \\
Transformer-XL~\cite{dai2019transformer} & V (ResNet-200) + F %
& 26.4 & 14.8 & 21.7 & 15.1 \\
MART~\cite{lei2020mart} & V (ResNet-200) + F %
& 35.7 & 15.9 & 23.4 & 15.7 \\
GVDSup~\cite{zhou2019grounded} & V (ResNet-101) + F + O %
& --- & --- & 22.9 & 16.4 \\
AdvInf~\cite{park2019adversarial} & V (ResNet-101) + F + O %
& --- & --- & 21.0 & 16.6 \\
PDVC~\cite{wang2021end} & V + F (TSN)
& --- & --- & 27.3 & 15.9 \\
\hline
\textit{With Learnt Proposals} & & & \\
MFT~\cite{xiong2018move} & V + F (TSN)
& --- & --- & 19.1 & 14.7 \\
PDVC~\cite{wang2021end} & V + F (TSN)
& --- & --- & 20.5 & 15.8 \\
PDVC~\cite{wang2021end}$^\dag$ & V (CLIP)
& --- & --- & 23.6 & 15.9 \\
\model{} (Ours) & V (CLIP)
& \textbf{50.1} & \textbf{24.0} & \textbf{28.0} & \textbf{17.0} \\
\bottomrule
\end{tabular}
}
\end{center}
\vspace{-0.6cm}
\caption{\small Comparison to the SoTA for video paragraph captioning.
$^\dag$ Results of our experiments using the official codebase. V/F/O refers to visual/flow/object features.}  
\vspace{-0.4cm}
\label{table:sotapara}
\end{table}

\begin{table}[t]
\begin{center}
\setlength\tabcolsep{1pt}
\resizebox{1.\linewidth}{!}{
\begin{tabular}{l|cc|cc|cc}
\toprule
\multirow{2}{*}{Method} & \multirow{2}{*}{\makecell{\small{Trained} \\ \small{Parameters}}} & \multirow{2}{*}{\makecell{\small{Pretraining} \\ \small{Data}}}
& \multicolumn{2}{c|}{MSR-VTT (test)}
& \multicolumn{2}{c}{MSVD (test)} \\
& & & \small{C} & \small{M}
& \small{C} & \small{M} \\
\midrule
ORG-TRL~\cite{zhang2020object} & --- & $\emptyset$ & 50.9 & 28.8 & 95.2 & 36.4 \\
SwinBERT~\cite{lin2022swinbert} & 229M & $\emptyset$ & 53.8 & 29.9
& 120.6 & 41.3 \\
\model{} (Ours) & 314M & $\emptyset$
& 57.2 & 30.0
& 120.3 & 41.4 \\
\hline
MV-GPT~\cite{seo2022end} & 354M & HowTo100M
& 60.0 & 29.9$^*$
& --- & --- \\
\model{} (Ours) & 314M & HowTo100M
& 61.5 & 30.4 
& 140.6 & 44.5 \\
\hline
\model{} (Ours) & 314M & YT-Temporal-1B
& \textbf{64.6} & \textbf{30.8}
& \textbf{146.2} & \textbf{45.3} \\
\bottomrule
\end{tabular}
}
\end{center}
\vspace{-0.6cm}
\caption{\small Comparison to the SoTA for video clip captioning.
* indicates results re-evaluated by the same evaluation toolkit.
}
\vspace{-0.4cm}
\label{table:sotaclip}
\end{table}

\noindent \textbf{Video paragraph captioning.} 
In Table~\ref{table:sotapara}, we compare our approach to state-of-the-art video paragraph captioning methods on the YouCook2 and ActivityNet Captions datasets.
\model{} outperforms all prior methods on both datasets, including the ones using ground-truth event boundary proposals at inference time~\cite{dai2019transformer, lei2020mart, zhou2018end, zhou2019grounded, wang2021end, park2019adversarial}, showing strong video paragraph captioning ability.

\noindent \textbf{Video clip captioning.} 
In Table~\ref{table:sotaclip}, we compare our approach to state-of-the-art video clip captioning methods on the MSR-VTT and MSVD datasets.
\model{} improves over prior methods in their respective pretraining data setting while using a comparable number of trained parameters.
We conclude that our pretrained \model{} model generalizes well to the standard video clip captioning setting.

\begin{table}[t]
\begin{center}
\setlength\tabcolsep{6pt}
\resizebox{1\linewidth}{!}{
\begin{tabular}{lc|ccc|ccc|ccc}
\toprule
& \multirow{2}{*}{Data}
& \multicolumn{3}{c|}{YouCook2}
& \multicolumn{3}{c|}{ViTT}
& \multicolumn{3}{c}{ActivityNet} \\
& & \small{S} & \small{C} & \small{M}
& \small{S} & \small{C} & \small{M} 
& \small{S} & \small{C} & \small{M} \\
\midrule
1. & 1\%
& 2.4 & 10.1 & 3.3
& 2.0 & 7.4 & 1.9
& 2.2 & 6.2 & 3.2 \\
2. & 10\%
& 3.8 & 18.4 & 5.2
& 10.7 & 28.6 & 6.0
& 4.3 & 20.0 & 6.1 \\
3. & 50\%
& 6.2 & 32.1 & 7.6
& 12.5 & 38.8 & 7.8
& 5.4 & 27.5 & 7.8 \\
4. & 100\%
& \textbf{7.9} & \textbf{47.1} & \textbf{9.3}
& \textbf{13.5} & \textbf{43.5} & \textbf{8.5}
& \textbf{5.8} & \textbf{30.1} & \textbf{8.5} \\
\bottomrule
\end{tabular}
}
\vspace{-0.3cm}
\caption{\small \textbf{Few-shot dense event captioning}, by finetuning \model{} using a small fraction of the downstream training dataset.}  
\label{table:fewshot}
\end{center}
\vspace{-1cm}
\end{table}

\subsection{Few-shot dense video captioning}\label{sec:fewshot}
To further evaluate the generalization capabilities of our pretrained \model{} model, we propose a new few-shot dense video captioning setting where we finetune \model{} using only a fraction of the downstream training dataset.
From Table~\ref{table:fewshot}, we observe important improvements when using 10\% compared to 1\% of training data (row 2 vs 1).
In Appendix Section~\ref{sec:fewshot2} we further show that pretraining is essential in this few-shot setting.

\vspace{-0.1cm}
\subsection{Qualitative examples}\label{sec:qualitative}
In Figure~\ref{fig:qualitative}, we show an example of dense event captioning predictions from \model{}.
This example shows that our model can predict meaningful event boundaries and captions, and that the predicted captions and boundaries differ considerably from the transcribed speech input (showing the importance of the visual tokens in the input).
More examples are provided in Appendix Section~\ref{sec:addquali}.

%% file: conclusion.tex
We introduced \model{}, a visual language model that performs dense video captioning by generating a single sequence of tokens including both text and time tokens given multi-modal inputs.
We showed that \model{} benefits from large-scale pretraining on unlabeled untrimmed narrated videos by leveraging transcribed speech sentences and corresponding temporal boundaries.
\model{} achieves state-of-the-art results on various dense event captioning datasets, as well as multiple video paragraph captioning and standard video clip captioning benchmarks.
Finally, we believe the sequence-to-sequence design of \model{} has the potential to be extended to a wide range of \textit{other} video tasks such as temporally-grounded video question answering~\cite{lei2018tvqa, li2021value, li2020hero} or temporal action localization~\cite{liu2022end, zhang2022actionformer, cheng2022tallformer}.

%% file: ack.tex
\mbox{}\vspace{-0.5cm}\\
\noindent 
{
\footnotesize{
\begin{spacing}{1.}
{\textbf{Acknowledgements.} The work was partially funded by a Google gift, the French government under management of Agence Nationale de la Recherche as part of the "Investissements d'avenir" program, reference ANR-19-P3IA-0001 (PRAIRIE 3IA Institute), the Louis Vuitton ENS Chair on Artificial Intelligence, the European Regional Development Fund under project IMPACT (reg.\ no.\ CZ.02.1.01/0.0/0.0/15 003/0000468).
We thank Anurag Arnab, Minsu Cho, Anja Hauth, Ashish Thapliyal, Bo Pang, Bryan Seybold and the entire Ganesha team for helpful discussions.
}
\end{spacing}
}}

%% file: appendix_arxiv.tex
In this Appendix, we present the following additional material:
\begin{itemize}
\item[\textit{(i)}] Additional qualitative examples of dense video captioning predictions (Section~\ref{sec:addquali}). 
\item[\textit{(ii)}] Additional information about our experimental setup (Section~\ref{sec:adddetails}); 
\item[\textit{(iii)}] Additional experimental results (Section~\ref{sec:addexperiments}), including an ablation on the importance of pretraining for few-shot dense video captioning (Section~\ref{sec:fewshot2}) and additional ablation studies in the standard fully-supervised dense video captioning setting (Section~\ref{sec:ablation2}).

\end{itemize}

\begin{figure*}[t]
\centering
\includegraphics[clip, trim=0mm 3.3cm 0mm 0mm, width=0.97\linewidth]{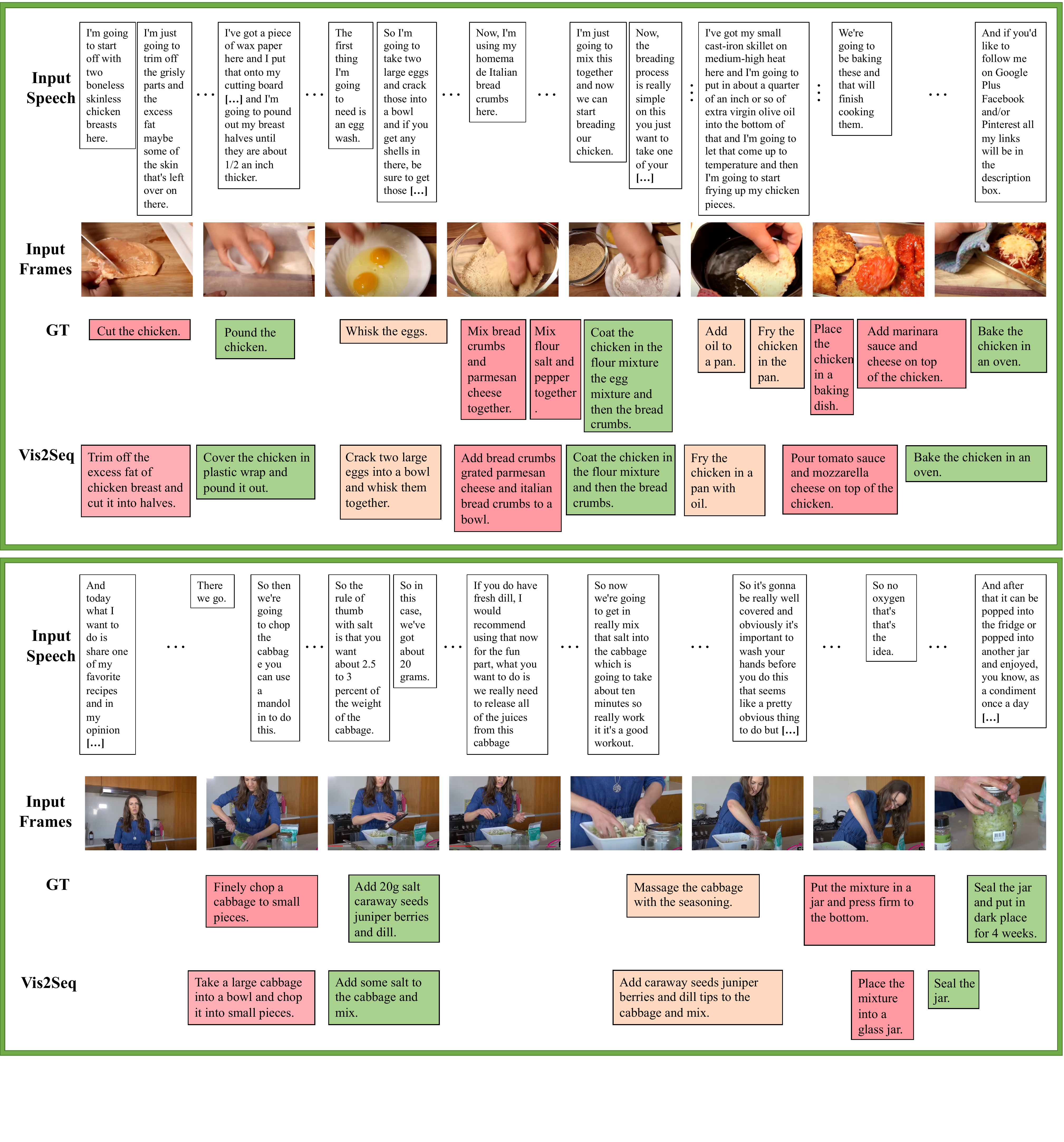}
\caption{\small 
Examples of dense event captioning predictions of \model{} on the validation set of YouCook2, compared with ground-truth.
}
\label{fig:qualitative2}
\end{figure*}

\begin{figure*}[t]
\centering
\includegraphics[clip, trim=0mm 1cm 0mm 0mm, width=0.97\linewidth]{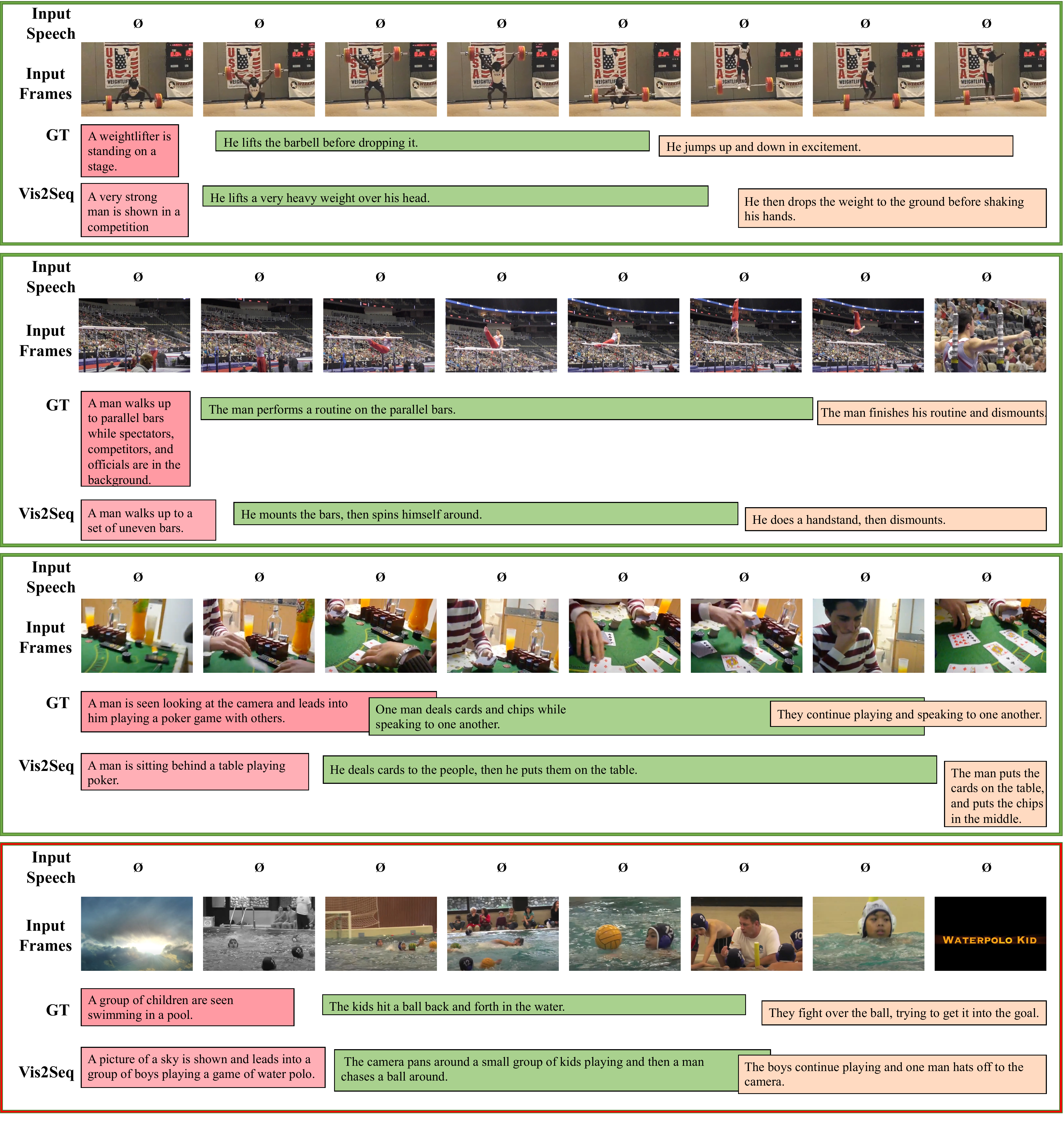}
\caption{\small 
Examples of dense event captioning predictions by \model{} on the validation set of ActivityNet Captions, compared with ground-truth.
The first three examples show successful predictions, while the last example illustrates a failure case where the model hallucinates events that are not visually grounded (`one man hats off to the camera`).
Note that in all of these videos, there is no transcribed speech.
}
\label{fig:qualitative3}
\end{figure*}

\section{Qualitative examples of dense video captioning predictions}\label{sec:addquali}

In Figure~\ref{fig:qualitative}, we show qualitative results of dense event captioning by our \model{} model.
Here in Figures~\ref{fig:qualitative2} and~\ref{fig:qualitative3} 
, we show additional results on examples from the YouCook2 and ActivityNet Captions datasets.
These results show that \model{} can predict meaningful dense captions and event boundaries in diverse scenarios, with or without transcribed speech input, \eg series of instructions in cooking recipes (Figure~\ref{fig:qualitative2}) or actions in human sports or leisure activities (first three examples in Figure~\ref{fig:qualitative3}).
The last example in Figure~\ref{fig:qualitative3} illustrates a failure case where the model hallucinates events that are not visually grounded such as `one man hats off to the camera`.

\section{Experimental setup}\label{sec:adddetails}

In this section, we complement the information provided in Section~\ref{sec:setup} about the datasets we use (Section~\ref{sec:adddatasets}). We also give additional implementation details (Section~\ref{sec:addimplem}).

\subsection{Datasets}\label{sec:adddatasets}

\noindent \textbf{YT-Temporal-1B}~\cite{zellers2022merlot} consists of 18.821M unlabeled narrated videos covering about 150 years of video content for pretraining.
Compared with HowTo100M~\cite{miech19howto100m}, this dataset was created to cover a wider range of domains and not only instructional videos.

\smallskip
\noindent \textbf{HowTo100M}~\cite{miech19howto100m} consists of 1.221M unlabeled narrated instructional videos covering about 15 years of video content for pretraining.

\smallskip
\noindent \textbf{YouCook2}~\cite{youcook2} has 1,790 untrimmed videos of cooking procedures.
On average, each video lasts 320s and is annotated with 7.7 temporally-localized imperative sentences.
The dataset is split into 1,333 videos for training and 457 videos for validation.

\smallskip
\noindent \textbf{ViTT}~\cite{huang2020multimodal} consists of 7,672 untrimmed instructional videos from the YouTube-8M dataset~\cite{abu2016youtube}.
Compared to YouCook2, ViTT was created to better reflect the distribution of instructional videos in the wild.
On average, each video lasts 250s and is annotated with 7.1 temporally-localized short tags.
The dataset is split into 5,476, 1,102 and 1,094 videos for training, validation and testing, respectively.
Videos in the validation and test sets are provided with multiple sets of dense event captioning annotations.
Following~\cite{huang2020multimodal}, we treat each set of annotations as a single example during evaluation and discard videos with more than 3 sets of annotations.

\smallskip
\noindent \textbf{ActivityNet-Captions}~\cite{krishna2017dense} contains 14,934 untrimmed videos of various human activities.
Different from YouCook2 and ViTT where most videos contain transcribed speech content, we find that 68\% of videos in ActivityNet Captions do not have transcribed narration.
On average, each video lasts 120s and is annotated with 3.7 temporally-localized sentences. 
The dataset is split into 10,009 and 4,925 videos for training and validation, respectively.
Videos in the validation set are provided with two sets of dense video captioning annotations.
Following prior work~\cite{wang2021end}, we use both sets of annotations for evaluation, by computing the average of the scores over each set for SODA\_c and by using the standard evaluation tool~\cite{krishna2017dense} for all other dense event captioning metrics.
For video paragraph captioning, we follow~\cite{wang2021end} and report results on the 'val-ae' split that includes 2,460 videos~\cite{zhou2019grounded, lei2020mart}.

\smallskip
\noindent \textbf{MSR-VTT}~\cite{xu16msrvtt} consists of 10,000 open domain video clips.
The duration of each video clip is between 10 and 30 seconds. 20 natural language descriptions are manually annotated for each clip.
The dataset is split into 6,513, 497 and 2,990 videos for training, validation and testing, respectively.

\smallskip
\noindent \textbf{MSVD}~\cite{chen2011collecting} consists of 1,970 open domain video clips.
The duration of each video clip is between 10 and 30 seconds. Each video clip has roughly 40 manually annotated captions.
The dataset is split into 1,200, 100 and 670 videos for training, validation and testing, respectively.

\subsection{Implementation details}\label{sec:addimplem}

\paragraph{Architecture.} 
The visual temporal transformer encoder $f^t$, the text encoder $g^t$ and the text decoder $h^t$ all have 12 layers, 12 heads, embedding dimension 768, and MLP hidden dimension of 2048.
The text encoder and decoder sequences are truncated or padded to $L=S=1000$ tokens during pretraining, and $S=1000$ and $L=256$ tokens during finetuning.
At inference, we use beam search decoding where we track the top 4 sequences and apply a length normalization of 0.6.

\paragraph{Training.} 
We use the Adam optimizer~\cite{kingma15adam} with $\beta=(0.9, 0.999)$ and no weight decay.
During pretraining, we use a learning rate of $1e^{-4}$, warming it up linearly (from 0) for the first 1000 iterations, and keeping it constant for the remaining iterations. 
During finetuning, we use a learning rate of $3e{-4}$, warming it up linearly (from 0) for the first 10\% of iterations, followed by a cosine decay (down to 0) for the remaining 90\%. 
During finetuning, we use a batch size of 32 videos split on 16 TPU v4 chips.
We finetune for 40 epochs on YouCook2, 20 epochs on ActivityNet Captions and ViTT, 5 epochs on MSR-VTT and 10 epochs on MSVD.
We clip the maximum norm of the gradient to 0.1 during pretraining, and 1 during finetuning.
For data augmentation, we use random temporal cropping.
For regularization, we use label smoothing~\cite{szegedy2016rethinking} with value 0.1 and dropout~\cite{srivastava2014dropout} with probability 0.1.

\section{Experiments}\label{sec:addexperiments}

In this section, we provide additional experiments that complement the results presented in Section~\ref{sec:experiments}.
We first show the importance of pretraining in our proposed few-shot setting in Section~\ref{sec:fewshot2}.
Then we provide additional ablation studies in the standard fully-supervised setting in Section~\ref{sec:ablation2}, where we ablate various factors including pretraining on long narrated videos,
the pretraining dataset and the size of the visual backbone, the time tokenization process and the number of time tokens, the sequence construction process, the temporal positional embeddings and the initialization of the language model.

\begin{table}[t]
\begin{center}
\setlength\tabcolsep{4pt}
\resizebox{1\linewidth}{!}{
\begin{tabular}{lcc|ccc|ccc|ccc}
\toprule
& \multirow{2}{*}{Data}
& \multirow{2}{*}{Pretrain}
& \multicolumn{3}{c|}{YouCook2}
& \multicolumn{3}{c|}{ViTT}
& \multicolumn{3}{c}{ActivityNet} \\
& & & \small{S} & \small{C} & \small{M}
& \small{S} & \small{C} & \small{M} 
& \small{S} & \small{C} & \small{M} \\
\midrule
1. & 1\% & \xmark
& 0.0 & 0.0 & 0.0
& 0.0 & 0.0 & 0.0
& 0.0 & 0.0 & 0.1 \\
2. & 1\% & \cmark
& 2.4 & 10.1 & 3.3
& 2.0 & 7.4 & 1.9
& 2.2 & 6.2 & 3.2 \\
3. & 10\% & \xmark
& 0.1 & 0.0 & 0.2
& 3.3 & 0.4 & 3.3
& 3.4 & 11.9 & 4.6 \\
4. & 10\% & \cmark
& 3.8 & 18.4 & 5.2
& 10.7 & 28.6 & 6.0
& 4.3 & 20.0 & 6.1 \\
5. & 50\% & \xmark
& 1.8 & 8.5 & 2.4
& 6.5 & 18.7 & 3.9
& 4.6 & 13.1 & 6.3 \\
6. & 50\% & \cmark
& 6.2 & 32.1 & 7.6
& 12.5 & 38.8 & 7.8
& 5.4 & 27.5 & 7.8 \\
7. & 100\% & \xmark
& 4.0 & 18.0 & 4.6
& 7.9 & 21.2 & 6.2
& 5.4 & 18.8 & 7.1 \\ 
8. & 100\% & \cmark
& \textbf{7.9} & \textbf{47.1} & \textbf{9.3}
& \textbf{13.5} & \textbf{43.5} & \textbf{8.5}
& \textbf{5.8} & \textbf{30.1} & \textbf{8.5} \\
\bottomrule
\end{tabular}
}
\vspace{-0.2cm}
\caption{\small \textbf{Impact of our pretraining on few-shot dense event captioning}, by finetuning \model{} using a small fraction of the downstream training dataset.}  
\label{table:fewshot2}
\end{center}
\vspace{-0.3cm}
\end{table}

\subsection{Importance of pretraining in few-shot settings}\label{sec:fewshot2}
In Section~\ref{sec:ablation}, we show the benefits of our pretraining method in the fully-supervised setting, \ie when using 100\% of the downstream training dataset.
In Table~\ref{table:fewshot2}, we further show that our pretraining method has a considerable importance in the few-shot setting defined in Section~\ref{sec:fewshot}, \ie when using a smaller fraction of the downstream training dataset. 
In particular, our pretraining method enables our \model{} model to have a non zero performance when using only 1\% of the downstream training dataset (rows 1 and 2).

\begin{table}[t]
\centering
\vspace{-0pt}
\begin{center}
\setlength\tabcolsep{6pt}
\resizebox{.85\linewidth}{!}{
\begin{tabular}{cc|ccc|ccc}
\toprule
& \multirow{2}{*}{\makecell{\small{Max number} \\ \small{of narrations}}} & \multicolumn{3}{c|}{YouCook2} & \multicolumn{3}{c}{ActivityNet} \\ & & \small{S} & \small{C} & \small{F1} & \small{S} & \small{C} & \small{F1} \\
\midrule
1. & \textit{No pretraining}
& 4.0 & 18.0 & 18.1 
& 5.4 & 18.8 & 49.2 \\ 
2. & 1
& 6.0 & 32.1 & 22.1
& 5.1 & 22.9 & 48.1 \\ 
3. & 10
& 6.5 & 34.6 & 23.6
& 5.4 & 27.1 & 50.3 \\ 
4. & $\infty$
& \textbf{7.9} & \textbf{47.1} & \textbf{27.3} 
& \textbf{5.8} & \textbf{30.1} & \textbf{52.4} \\ 
\bottomrule
\end{tabular}}
\vspace{-0.2cm}
\caption{\small \textbf{Ablation showing the importance of pretraining on long narrated videos}, by varying the maximum number of narration sentences that a randomly cropped video can cover.
$\infty$ means the cropping is unrestricted and can sample arbitrarily long videos.}
\vspace{-0.3cm}
\label{table:long}
\end{center}
\end{table}

\begin{table}[t]
\begin{center}
\setlength\tabcolsep{6pt}
\resizebox{1\linewidth}{!}{
\begin{tabular}{lcc|ccc|ccc}
\toprule
& \multirow{2}{*}{Pretraining Data}
& \multirow{2}{*}{Model}
& \multicolumn{3}{c|}{YouCook2}
& \multicolumn{3}{c}{ActivityNet} \\
& & & \small{S} & \small{C} & \small{F1}
& \small{S} & \small{C} & \small{F1} \\
\midrule
1. & ImageNet & ViT-B/16
& 6.6 & 40.2 & 24.3
& 4.5 & 17.2 & 49.3 \\
2. & CLIP & ViT-B/16
& 7.7 & 46.3 & 26.5
& 5.6 & 28.4 & 51.7 \\
3. & CLIP & ViT-L/14
& \textbf{7.9} & \textbf{47.1} & \textbf{27.3} 
& \textbf{5.8} & \textbf{30.1} & \textbf{52.4} \\

\bottomrule
\end{tabular}
}
\vspace{-0.2cm}
\caption{\small \textbf{Ablation on the pretraining data and model size of the visual backbone $f^s$.}}  
\label{table:backbone}
\end{center}
\vspace{-0.3cm}
\end{table}

\subsection{Additional ablation studies}\label{sec:ablation2}
We here complement ablation studies reported in Section~\ref{sec:ablation}, using the same default settings, evaluation metrics and downstream datasets.

\paragraph{Pretraining on long narrated videos.}
In Table~\ref{table:pretraining}, we show the benefits of pretraining on untrimmed videos in comparison with the standard practice of pretraining on short, trimmed, video-speech segments~\cite{huang2020multimodal, luo2020univilm, seo2022end}. 
In Table~\ref{table:long}, we further evaluate the importance of sampling long narrated videos during pretraining.
By default, at each training iteration, we randomly temporally crop each narrated video without constraints, resulting in a video that can span over hundreds of transcribed speech sentences.
We here evaluate a baseline that constrains this cropping process such that the cropped video only spans over a given maximum number of narration sentences.
Even with a maximum of 10 narration sentences, this baseline significantly underperforms our model trained in default settings where we sample longer untrimmed narrated videos (rows 1, 2 and 3).
This demonstrates that our model benefits from pretraining on long narrated videos. 

\paragraph{Visual features.}
In Table~\ref{table:scale}, we show the benefits of scaling up the size of the pretraining dataset of narrated videos and the size of the language model. 
In Table~\ref{table:backbone}, we further analyze the importance of the pretraining dataset and size of the visual backbone $f^s$.
We find that CLIP pretraining~\cite{radford2021learning} considerably improves over ImageNet pretraining~\cite{steiner2021train} with the same ViT-B/16 visual backbone model (row 2 vs 1).
Furthermore, scaling up the visual backbone size from ViT-B/16 to ViT-L/14 brings additional improvements (row 3 vs 2).

\begin{table}[t]
\begin{center}
\setlength\tabcolsep{6pt}
\resizebox{1\linewidth}{!}{
\begin{tabular}{lcc|ccc|ccc}
\toprule
& \multirow{2}{*}{Tokenization}
& \multirow{2}{*}{$N$}
& \multicolumn{3}{c|}{YouCook2}
& \multicolumn{3}{c}{ActivityNet} \\
& & & \small{S} & \small{C} & \small{F1}
& \small{S} & \small{C} & \small{F1} \\
\midrule
1. & Absolute & 20
& 0.3 & 0.2 & 0.9
& 3.2 & 23.0 & 23.1 \\
2. & Absolute & 100
& 3.5 & 25.7 & 12.0
& 4.8 & 25.5 & 41.5 \\
3. & Absolute & 500
& \textbf{7.9} & 39.8 & 24.3
& 5.4 & 28.1 & 48.6 \\
4. & Relative & 20
& 7.2 & 39.6 & 23.7
& 5.6 & 29.0 & 49.4 \\
5. & Relative & 100
& \textbf{7.9} & \textbf{47.1} & \textbf{27.3} 
& \textbf{5.8} & \textbf{30.1} & 52.4 \\
6. & Relative & 500
& 7.2 & 40.0 & 25.0
& 5.7 & 28.6 & \textbf{52.5} \\
\bottomrule
\end{tabular}
}
\vspace{-0.2cm}
\caption{\small \textbf{Ablation on time tokenization (relative or absolute) and the number of time tokens $N$.}}  
\label{table:time}
\end{center}
\vspace{-0.3cm}
\end{table}

\begin{table}[t]
\centering
\vspace{-0pt}
\begin{center}
\setlength\tabcolsep{4pt}
\resizebox{1.\linewidth}{!}{
\begin{tabular}{ccc|ccc|ccc}
\toprule
& \multirow{2}{*}{\makecell{\small{Dot symbol} \\ \small{between segments}}} & \multirow{2}{*}{\makecell{\small{Time tokens} \\ \small{Position}}} & \multicolumn{3}{c|}{YouCook2} & \multicolumn{3}{c}{ActivityNet} \\ & & & \small{S} & \small{C} & \small{F1} & \small{S} & \small{C} & \small{F1} \\
\midrule
1. & \xmark & \textit{After text}
& 7.9 & 48.3 & 26.7
& 5.6 & 29.8 & 51.1 \\ 
2. & \cmark & \textit{After text}
& \textbf{8.3} & \textbf{50.9} & 26.2
& 5.7 & \textbf{30.4} & 51.8 \\ 
3. & \xmark & \textit{Before text}
& 8.0 & 50.0 & \textbf{27.3}
& 5.6 & 28.2 & 50.7 \\ 
4. & \cmark & \textit{Before text}
& 7.9 & 47.1 & \textbf{27.3} 
& \textbf{5.8} & 30.1 & \textbf{52.4} \\ 
\bottomrule
\end{tabular}}
\vspace{-0.2cm}
\caption{\small \textbf{Ablation on the sequence construction process.}}
\vspace{-0.3cm}
\label{table:sequence}
\end{center}
\end{table}

\paragraph{Time tokenization and number of time tokens.}
In Table~\ref{table:time}, we further ablate the time tokenization process presented in Section~\ref{sec:model}.
Our default time tokens represent relative timestamps in a video, as we quantize a video of duration $T$ into $N$ equally-spaced timestamps.
Another possibility is to use time tokens that represent absolute timestamps in the video, \ie the k-th token represents the k-th second in the video.
For both these variants, we vary the number of time tokens $N$. 
For the relative time tokens, increasing $N$ makes the quantization more fine-grained but also spreads the data into more time tokens. 
On the other hand, for the absolute time tokens, increasing $N$ increases the video duration that the time tokens can cover.
We find that the best dense video captioning results are obtained with the relative time tokens and $N=100$ time tokens (row 5).

\paragraph{Sequence construction.}
In Table~\ref{table:sequence}, we further ablate the sequence construction process presented in Section~\ref{sec:model}.
Our default sequence inserts the start and end time tokens of each segment before its corresponding text sentence.
Another possibility is to insert time tokens after each corresponding text sentence.
We find that both variants achieve similar results (rows 2 and 4), with the default sequence (row 4) resulting in slightly higher event localization performance (F1 Score) but slightly lower dense captioning results overall.
Furthermore, we observe that the dot symbols indicating the separation between different events have low importance (rows 1 and 2, rows 3 and 4).

\paragraph{Temporal positional embeddings.}
In Table~\ref{table:pretraining}, we show that time tokens in the speech sequence provide temporal information about the speech transcript to our model.
In Table~\ref{table:temporal}, we also evaluate the importance of the temporal positional embeddings which communicate temporal information from the visual stream to our model.
We find that these temporal embeddings are beneficial (row 2 vs 1).

\begin{table}[t]
\centering
\vspace{-0pt}
\begin{center}
\setlength\tabcolsep{6pt}
\resizebox{.85\linewidth}{!}{
\begin{tabular}{cc|ccc|ccc}
\toprule
& \multirow{2}{*}{\makecell{\small{Temporal} \\ \small{embeddings}}} & \multicolumn{3}{c|}{YouCook2} & \multicolumn{3}{c}{ActivityNet} \\ & & \small{S} & \small{C} & \small{F1} & \small{S} & \small{C} & \small{F1} \\
\midrule
1. & \xmark
& 6.8 & 42.0 & 24.9
& 5.3 & 27.0 & 50.6 \\ 
2. & \cmark
& \textbf{7.9} & \textbf{47.1} & \textbf{27.3} 
& \textbf{5.8} & \textbf{30.1} & \textbf{52.4} \\ 
\bottomrule
\end{tabular}}
\vspace{-0.2cm}
\caption{\small \textbf{Ablation on the temporal positional embeddings.}}
\vspace{-0.3cm}
\label{table:temporal}
\end{center}
\end{table}

\begin{table}[t]
\centering
\vspace{-0pt}
\begin{center}
\setlength\tabcolsep{4pt}
\resizebox{1.\linewidth}{!}{
\begin{tabular}{ccc|ccc|ccc}
\toprule
& \multirow{2}{*}{\makecell{\small{Language Model} \\ \small{Initialization}}} & \multirow{2}{*}{\makecell{\small{Video-text} \\ \small{Pretraining}}} & \multicolumn{3}{c|}{YouCook2} & \multicolumn{3}{c}{ActivityNet} \\ & & & \small{S} & \small{C} & \small{F1} & \small{S} & \small{C} & \small{F1} \\
\midrule
1. & \xmark & \xmark
& 0.9 & 4.2 & 7.6
& 4.3 & 23.7 & 41.2 \\ 
2. & \cmark & \xmark
& 4.0 & 18.0 & 18.1
& 5.4 & 18.8 & 49.2 \\ 
3. & \xmark & \cmark
& \textbf{8.8} & \textbf{51.3} & \textbf{28.4}
& 5.7 & 28.7 & 51.2 \\ 
4. & \cmark & \cmark 
& 7.9 & 47.1 & 27.3 
& \textbf{5.8} & \textbf{30.1} & \textbf{52.4} \\ 
\bottomrule
\end{tabular}}
\vspace{-0.2cm}
\caption{\small \textbf{Ablation on language model initialization and pretraining.}}
\vspace{-0.3cm}
\label{table:initialization}
\end{center}
\end{table}

\paragraph{Language model initialization and pretraining.}
In Table~\ref{table:scale}, we show the benefits of using T5-Base instead of T5-Small.
In Table~\ref{table:initialization}, we further investigate the importance of initializing the language model from weights pretrained on Web text.
Without pretraining on narrated videos, we find that text-only initialization is helpful (rows 1 and 2).
Interestingly, after pretraining on narrated videos, we find that text-only initialization has little importance (rows 3 and 4), as it slightly improves the performance on ActivityNet Captions while resulting in a slight drop of performance on YouCook2.
We believe that this may be because of the domain gap between Web text and the imperative-style dense captions in YouCook2, which are more similar to transcribed speech in YT-Temporal-1B.